\documentclass[twocolumn]{article}
\pdfoutput=1
\usepackage[utf8]{inputenc}
\usepackage{amsfonts}
\usepackage{amsmath}
\usepackage{amsthm}
\usepackage{authblk} % needed for affiliation
\usepackage{graphicx}
\newtheorem{definition}{Definition}

\title{Towards the Linear Algebra Based Taxonomy of XAI Explanations}
\author{Sven N{\~o}mm}
\affil{Department of Software Science, School of Information Technology, Tallinn University of Technology, (TalTech), Akadeemia tee 15 a, Tallinn, 12618, Estonia, E-mail: sven.nomm@taltech.ee}
\date{}

\begin{document}

\maketitle
\begin{abstract}
This paper proposes an alternative approach to the basic taxonomy of explanations produced by explainable artificial intelligence techniques. Methods of Explainable Artificial Intelligence (XAI) were developed to answer the question why a certain prediction or estimation was made, preferably in terms easy to understand by the human agent. XAI taxonomies proposed in the literature mainly concentrate their attention on distinguishing explanations with respect to involving the human agent, which makes it complicated to provide a more mathematical approach to distinguish and compare different explanations. This paper narrows its attention to the cases where the data set of interest belongs to $\mathbb{R} ^n$ and proposes a simple linear algebra-based taxonomy for local explanations. 

\end{abstract}


\section{Introduction}
The necessity to explain or interpret the decisions of artificial intelligence (AI) classifiers has led to the formation of the domain of XAI, according to \cite{schwalbe2021xai} the abbreviation XAI was first proposed by \cite{van2004explainable}. On the machine learning side of the problem, it allows one to improve the goodness of the classifiers and find the limits of their applicability. At the same time, the widespread application of AI in life-critical areas such as medicine \cite{YANG202229} or power supply \cite{MACHLEV2022100169}, \cite{Meas_MDPI_XAI_2022} requires explanation not only for technical purposes, but also to win the trust of practitioners. The latest may be seen as an integral component of transparency in the application of AI techniques. 
Existing taxonomies, for example \cite{BARREDOARRIETA2020_InformationFusion_taxonomy} or \cite{schwalbe2021xai} are based on human-on-the-loop and human-in-the-loop paradigms. This leads to certain bias towards human-readable results, whereas the formal mathematical part of the problem is becoming increasingly forgotten. However, the problem of explaining classifier decisions is closely related to the problem of feature selection in statistical machine learning, which justifies alternative, linear algebra-based definitions and taxonomy. Without undermining the importance of the properties that describe the human perception of XAI output, these properties are beyond the scope of this research. In the best knowledge of the author up to the present moment there is no taxonomy which is purely based on the linear-algebra approach.

In the present contribution, we propose to distinguish between the decision {\em trace}, {\em approximation}, and {\em interpretation}. The first formal (linear algebra-based) definition of these notions will be provided and illustrated with synthetic examples. Then, their relations with similar notions in existing taxonomies will be discussed. This leads to the second goal of the paper, which is to spark the discussion leading to mathematical formalisation of the existing "human-centric" notions and definitions. 

The remainder of the paper is organised as follows. Section \ref{sec:background} provides the necessary information about the principles of XAI and relates the current work to previously published results. The proposed definitions are given in Section \ref{sec:definitions}. Section \ref{sec:examples} hosts a numeric example that illustrates the proposed taxonomy. The short discussion in Section \ref{sec:discussion} is followed by the conclusive remarks stated in the last section. 

\section{Background}\label{sec:background}
Without undermining the importance of previous work such as \cite{van2004explainable}, \cite{NIPS1995_45f31d16} devoted to explaining classifier decisions, one may refer to \cite{Ribeiro_2017} and \cite{lundberg2017unified} as works resulting in popular software packages LIME and SHAP, respectively, which sparked greater interest in XAI. In the case of a numerical problem, for each classified point, most XAI packages return the set of inequalities that describe the position of this point with respect to the decision boundary. For each feature, there is at least one inequality that describes the value of the feature with respect to the boundaries separating different classes. These inequalities are ordered or accompanied by the weight referring to the importance of the feature in making the decision (class label prediction). 

Although \cite{BARREDOARRIETA2020_InformationFusion_taxonomy}, \cite{Bruckert_FrontiersAI_2020_AIMed}, and \cite{murdoch2019interpretable} mainly cite and use human-centric definitions and taxonomies, some of their results will be used to relate the concepts discussed in this document to those available in the literature. Existing results mainly devoted to complex taxonomies distinguishing between such notions as {\em understanding} \cite{Bruckert_FrontiersAI_2020_AIMed}, {\em explicability}, {\em transparency}, etc. Among these long lists of notions, {\em explanation} and {\em interpretation} will be discussed in the present paper and compared with the existing definitions later in Section \ref{sec:discussion}.

\section{Proposed definitions}\label{sec:definitions}
Consider the two-class classification problem.  Let $D \subset  \mathbb{R} ^m$ be a $m$ dimensional data set. Denote $\{ \lambda _1, \lambda _2 \ldots , \lambda _m\}$ as the basis for $\mathbb{R} ^m$.  Following the standard machine learning workflow, we can assume the splitting of the data set $D$ into training validation and testing subsets $D = D_{\rm{train}} \cup D_{\rm{valid}} \cup D_{\rm{test}}$. Let $X \subseteq \{ \lambda _1, \lambda _2 \ldots , \lambda _m\}$ denote the feature set used to train the classifier $C$ and $\hat{y} = C(x_i) $ the estimation of the class label given by $C$ to the point $x_i \in \rm{proj}_X (D_{test})$ (the projection of the set $D_{\rm{test}}$ in the space spanned by the feature set $X$). 
\begin{definition}
 $\varepsilon $- decision boundary of the classifier $C$ is the set of points in $S$ such that the $\varepsilon$- neighbourhood of each point contains at least one point of each class.
\end{definition}
Note that the class label is predicted by the classifier $C$ which implies that the decision boundary is defined by the classifier. From a practical point of view, to construct the decision boundary, it may be wiser to draw the set $D_{\rm{synth}} \subset \mathbb{R}^m$ from the same distribution as the initial data set $D$. And apply the classifier to each point of the set $D_{\rm{synth}} \setminus D_{\rm{train}}$. 
\begin{figure}[h]\label{fig:decision_boundary}
    \includegraphics[width=0.49\textwidth]{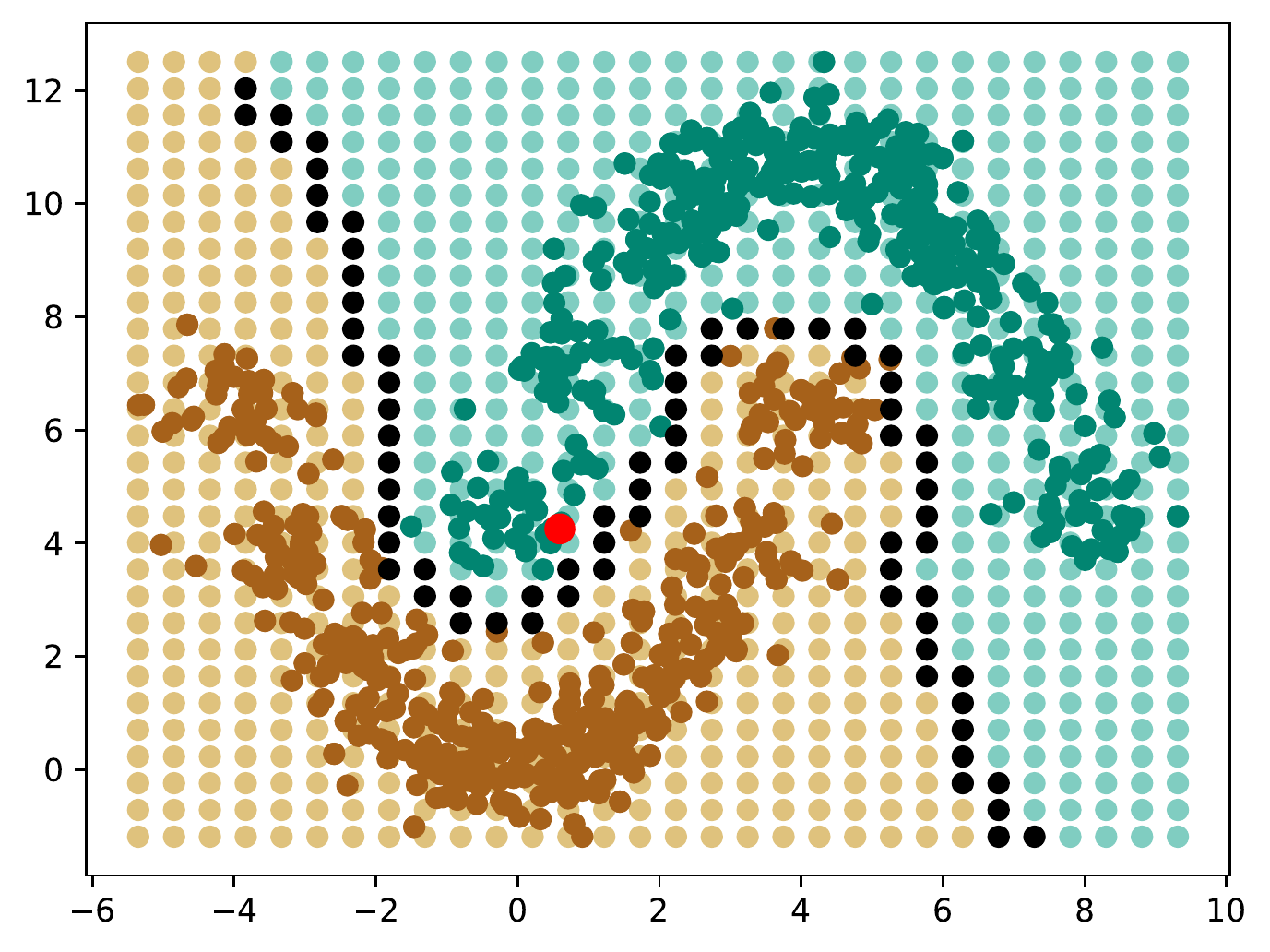}\\
    \caption{Decision boundary for the two class {\em half moons} dataset.}
\end{figure}
In Figure \ref{fig:decision_boundary} the dark brown and dark green points represent two classes of the well-known {\em half moons} dataset, used to train the $k$NN classifier. The light brown and light green points represent the classified synthetic data set, and the black points depict the decision boundary. The scatter-type plot was chosen purposefully to emphasise that the decision boundary is a set of points.   

\begin{definition}
The prediction trace (or trace of $\hat{y_i} = C(x_i) $) is the set of at least $m$ inequalities that explain the position of $x_i$ with respect to the decision boundary in the coordinate system defined by $X$. Whereas at least one inequality should correspond to each feature. Each inequality can be complemented by the weight that describes its importance in making the prediction of the labels. 
\end{definition}
Note that the decision boundaries may require more than one inequality for each feature. For example, in Figure \ref{fig:decision_boundary} for the point marked by the red colour, two inequalities are required to explain the position of the point, for the feature represented by the horizontal axis. Generally speaking, all features of the feature set should be presented in the trace. The prediction trace should not be confused with the trace points on the Pareto frontier \cite{kamani2021pareto}. Let us consider common classifiers commonly cited in the area of statistical learning \cite{hastie}. 

\begin{itemize}
	 \item Decision trees naturally produce the set of inequalities. The number of inequalities in this case may differ from $m$.  If the number of inequalities is smaller than $m$ and some features are not used, then the set can be complemented by inequalities of the form $ - \infty < x_i < \infty$. The order of the conditions provides a natural way to describe their importance for decision-making. 
	 \item Support vector machine SVMs may explicitly provide a linear or nonlinear equation of the decision boundary in more than one variable. in this case, one may convert it into a system of parametric inequalities.
	 \item $k$- nearest neighbours does not produce any inequalities.
	 \item logistic regression naturally produces a decision boundary, whereas the position of each point may be explicitly explained with respect to this decision boundary.
\end{itemize}

Feature selection techniques may not necessarily be optimised for the particular classifier, which may cause features with low impotence (low influence on the classification results). For example, in the case of decision trees, pruning branches that lead to a smaller information gain may cause some features to be unused. Together, this leads to the idea of explaining the classifier decision (label prediction) with a lower number of features. This may be seen as the projection of the initial data set into the subspace of $\mathbb{R}^m$ spanned by the subset $X_r$ of $X$.

\begin{definition}
The prediction approximation is the set of inequalities that describe the position of $\hat{y_i} = C(x_i)$ with respect to the decision boundary in the coordinate system defined by $X_r \subset X$. Whereas, each inequality can be complemented by the weight that describes its importance in predicting the label. 
\end{definition}
Note that only features belonging to $X_r$ may be presented in the approximation. An approximation of the prediction may be seen as an attempt to reduce the dimensionality of the prediction trace.\\

The idea of interpretation is to explain the prediction of the class given by the classifier $C$ in terms of the set of features that differs from $X$. This implies the necessity of mapping $M_I :\rm{proj} _{X}(D) \rightarrow \tilde{D}$ where $\tilde{D}$ is spanned by the set of features to be used in the approximation. An easy example are the features that span $S$ except $X$.
More formally, denote by $X_S$ the set of features that spans space $S$, then attempt to interpret the classification results given by $C$ in terms of features $X_S \setminus X$.

\begin{definition}\label{def:interpretation}
The interpretation of the prediction is the set of inequalities that explain the position of $M_I(x_i)$ with respect to the image of the decision boundary in the coordinate system defined by $M_I(X)$.  Whereas each inequality may be complemented by the weight describing its importance in making label prediction. 
\end{definition}
The weighting of inequalities with respect to their importance may be the most challenging task in this case.
Interpretation may be seen as mapping the trace or approximation to the most suitable coordinates for human understanding or simplifying the task, for example, converting the non-linear decision boundary into the linear one \cite{aggarwal2016data}. One may easily see that the approximation is one particular case of interpretation, where the mapping $M_I$ is a projection in a space of lower dimension. The definition \ref{def:interpretation} implies the existence of a mapping $M_I :\rm{proj}_{X} (D) \rightarrow \tilde{D}$ where $\tilde{D}$ is spanned by the set of features to be used in the approximation. Approximation can be applied in the areas of medicine where feature engineering techniques have resulted in unusual feature sets for medical professionals, such as analysis drawing and writing tests \cite{Nomm_at_al_Luria_analysis_ICMLA2018}. In such cases mapping the data to the space spanned by the feature that describes the tremor \cite{VALLA2022103551} makes a lot of sense with respect to making the results human-comprehensible.

\section{Examples}\label{sec:examples}
\begin{figure}[h]
    \includegraphics[width=0.49\textwidth]{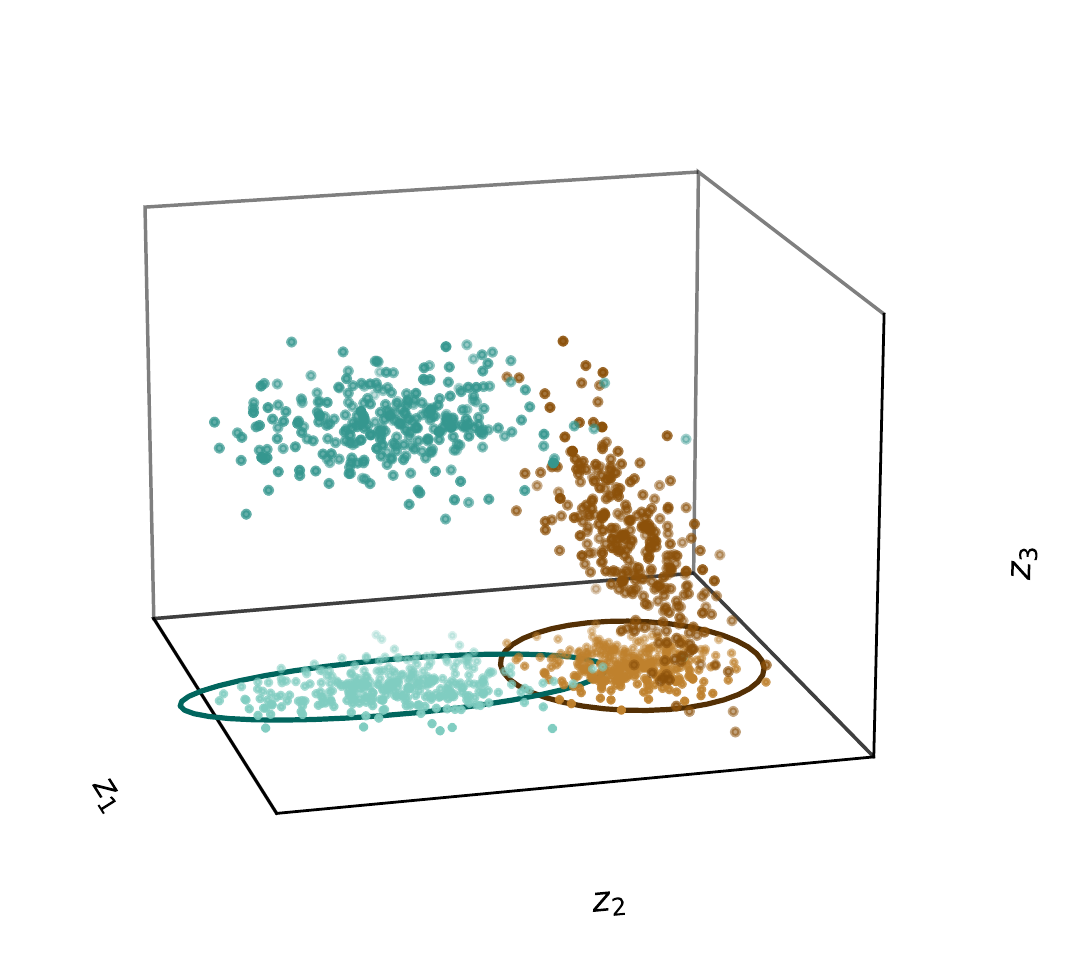}\\
    \caption{The case when first two dimensions are more informative.}\label{fig:elliptic}
\end{figure}

Decision {\em trace} is the most common type produced by XAI techniques for local explanation such as LIME \cite{Ribeiro_2017} or SHAP \cite{lundberg2017unified}.
To illustrate the decision approximation, let us consider the case where a classifier is trained on the dataset consisting of two partially overlapping elliptically shaped classes, as shown in Figure \ref{fig:elliptic}. 

In this case, filter models for feature selection \cite{aggarwal2016data} may not necessarily disregard feature $z$. At the same time, it may not necessarily be important to explain the classification decision, or its importance may be rather low. For this, one may prefer to explain ({\em approximate}) the decision of the classifier using only a projection on two-dimensional space.\\

\begin{figure}[h]
    \includegraphics[width=0.49\textwidth]{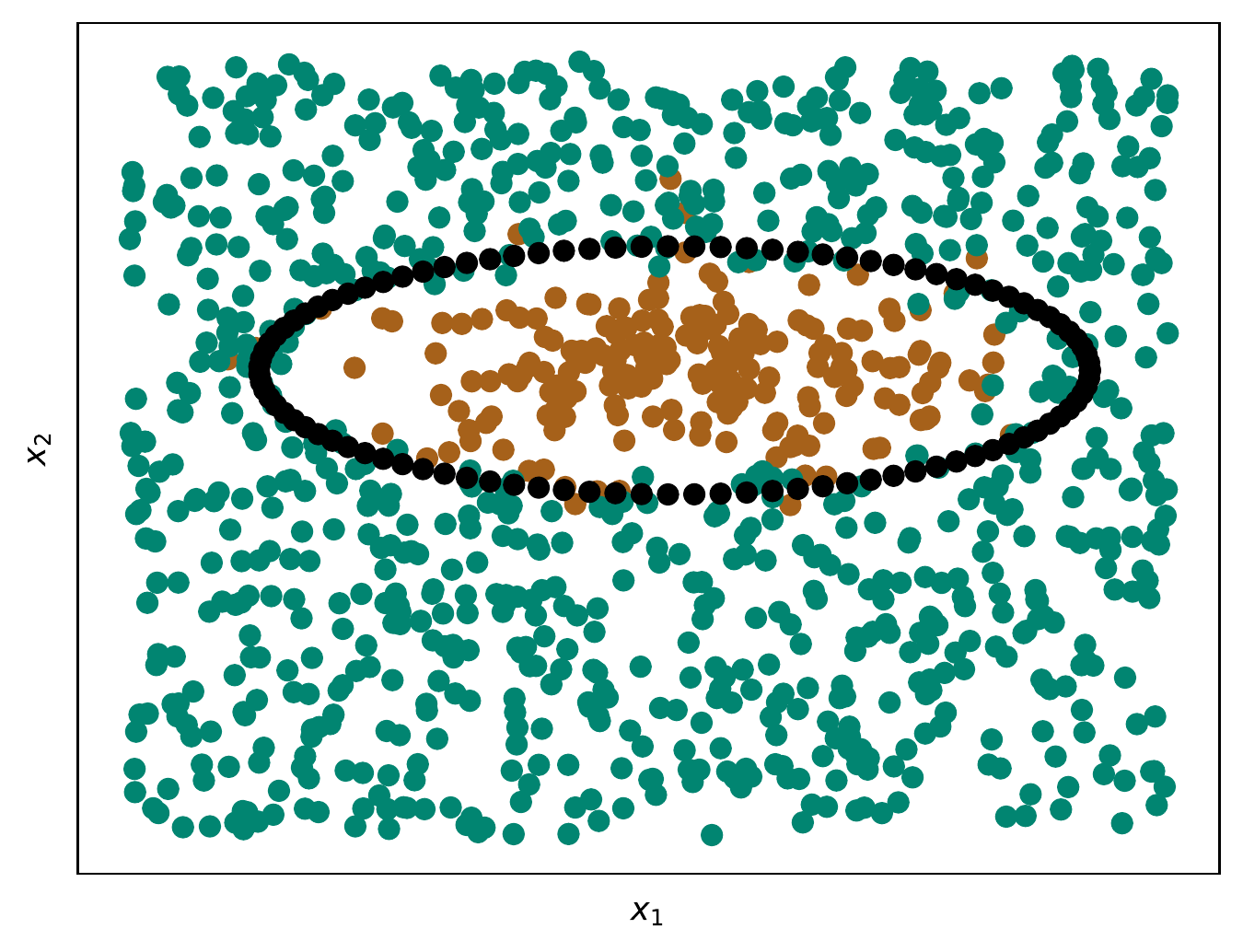}\\
    \caption{Case of ellipse as a decision boundary.}\label{fig:decision_boundary_ellipse}
\end{figure}

To illustrate decision {\em interpretation} let us use the example proposed by \cite{aggarwal2016data}. Assume elliptic decision boundary, depicted in Figure \ref{fig:decision_boundary_ellipse}. 

\begin{figure}[h]
    \includegraphics[width=0.49\textwidth]{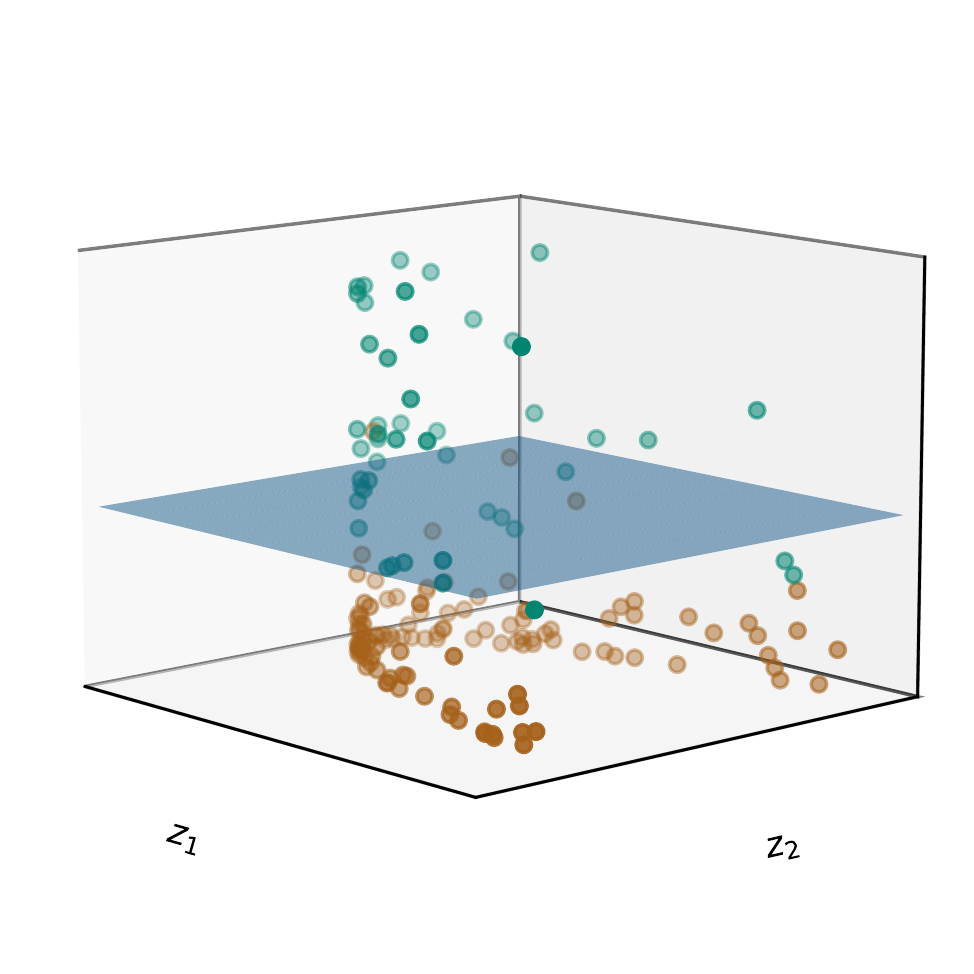}\\
    \caption{Decision boundary corresponding to the ellipse after projection into the space of higher dimension.}\label{fig:decision_boundary_plane}
\end{figure}

One may prefer to solve this problem by projecting it into a three-dimensional space with a simple variable change.
\begin{eqnarray*}
    z_1 &=& x_1 ^2 \\
    z_2 &=& z_2\\
    z_3 &=& x_2^2
\end{eqnarray*}
where $x_1$ corresponds to the horizontal axis, $x_2$ to the vertical one, we also assume that along $x_2$ the centre of the class is positioned at $0$. Being projected in a three-dimensional space, the decision boundary will become linear as depicted in Figure \ref{fig:decision_boundary_plane}.
Although model explainer will produce the trace of classification decision using the feature set of original problems, in some cases one may desire to interpret the decision using the other feature set. \\

Another possible example is as follows: Let $D$ be a data set in the six-dimensional space. Let three even features be uncorrelated, but each of them will be tightly linearly correlated to the corresponding odd feature. For the classifier $C$ trained only on features of even features, it can be attempted to explain the classification results using odd features instead. The applicability of such an approach may be questioned; but formally, such a procedure is correct.

\section{Discussion}\label{sec:discussion}
In \cite{Bruckert_FrontiersAI_2020_AIMed} the authors distinguish between the notions of explicability and explainable which most probably would be the closest analogues to decision tracing and interpreting. In addition, the definition of transparency given by \cite{paez2019pragmatic} may cover the cases of decision trace and decision approximation proposed in this paper. The notion of interpretability is probably the most difficult to match. In \cite{murdoch2019interpretable} interpretability refers more to the ability to capture relationships that the model has learnt. On the other hand, the interpretability described by \cite{BARREDOARRIETA2020_InformationFusion_taxonomy} is completely similar to that proposed in this paper. In addition, \cite{BARREDOARRIETA2020_InformationFusion_taxonomy} defines transparency in a way similar to \cite{paez2019pragmatic}. Although most of the cited results lack formal mathematical definitions, it is clear that the taxonomy proposed in this paper may be related to existing once.

\section{Conclusions}
The present paper proposes a simple linear algebra-based taxonomy of the XAI explanations. 
Although in first view it differs significantly from those proposed before, it was clearly demonstrated that proposed notions may be related to the existing once.

\section*{Acknowledgements}
This work in the project "ICT programme" was supported by the European Union through European Social Fund.
\bibliography{xai_trace_explain_interpret}
\end{document}